%% file: arxiv.tex
\documentclass[final]{cvpr}

\usepackage{times}
\usepackage{epsfig}
\usepackage{graphicx}
\usepackage{amsmath}
\usepackage{amssymb}
\usepackage{url}
\usepackage{multirow}
\usepackage{multicol}
\usepackage{color}
\definecolor{citecolor}{RGB}{119,185,0} 
\usepackage[pagebackref=true,breaklinks=true,letterpaper=true,colorlinks,citecolor=citecolor,bookmarks=false]{hyperref}

\usepackage[pagebackref=true,breaklinks=true,colorlinks,bookmarks=false]{hyperref}
\usepackage{enumitem}

\usepackage{comment}

\usepackage{xparse}
\ExplSyntaxOn
\NewDocumentCommand{\mref}{m}{\quinn_mref:n {#1}}
\seq_new:N \l_quinn_mref_seq
\cs_new:Npn \quinn_mref:n #1
 {
  \seq_set_split:Nnn \l_quinn_mref_seq { , } { #1 }
  \seq_pop_right:NN \l_quinn_mref_seq \l_tmpa_tl
  ( 
  \seq_map_inline:Nn \l_quinn_mref_seq
    { \ref{##1},\nobreakspace } 
  \exp_args:NV \ref \l_tmpa_tl 
  ) 
 }
\ExplSyntaxOff

\usepackage{xcolor}
\usepackage{colortbl}

\usepackage{amssymb}
\usepackage{pifont}
\newcommand{\cmark}{\ding{51}}%
\newcommand{\xmark}{\ding{55}}%

\newcommand\blfootnote[1]{%
  \begingroup
  \renewcommand\thefootnote{}\footnote{#1}%
  \addtocounter{footnote}{-1}%
  \endgroup
}

\def\cvprPaperID{3524} 
\def\confYear{CVPR 2021}

\begin{document}

\title{Self-Supervised Pillar Motion Learning for Autonomous Driving}

\author{
  Chenxu Luo${^{1,2}}$ \quad Xiaodong Yang${^1}$ \quad Alan Yuille${^2}$\\
  $^1$QCraft \quad $^2$Johns Hopkins University \\
}

\maketitle

\begin{abstract}
Autonomous driving can benefit from motion behavior comprehension when interacting with diverse traffic participants in highly dynamic environments. Recently, there has been a growing interest in estimating class-agnostic motion directly from point clouds. Current motion estimation methods usually require vast amount of annotated training data from self-driving scenes. However, manually labeling point clouds is notoriously difficult, error-prone and time-consuming. In this paper, we seek to answer the research question of whether the abundant unlabeled data collections can be utilized for accurate and efficient motion learning. To this end, we propose a learning framework that leverages free supervisory signals from point clouds and paired camera images to estimate motion purely via self-supervision. Our model involves a point cloud based structural consistency augmented with probabilistic motion masking as well as a cross-sensor motion regularization to realize the desired self-supervision. Experiments reveal that our approach performs competitively to supervised methods, and achieves the state-of-the-art result when combining our self-supervised model with supervised fine-tuning.

\end{abstract}

\section{Introduction}

Understanding\blfootnote{This work was done while C. Luo was interning at QCraft.} the motion of various traffic agents is crucial for self-driving vehicles to be able to safely operate in dynamic environments. Motion provides pivotal information to facilitate a variety of onboard modules ranging from detection, tracking, prediction to planning. A self-driving vehicle is typically equipped with multiple sensors, and the most commonly used one is LiDAR. How to represent and extract temporal motion from point clouds is therefore one of the fundamental research problems in autonomous driving~\cite{vector-net, flownet3d, motion-net}. This is however challenging in the sense that (1) there exist numerous agent categories and each category exhibits specific motion behavior; (2) point cloud is sparse and lacks of exact correspondence between sweeps; and (3) estimating process is required to meet tight runtime constraint and limited onboard computation.

\begin{figure}[t]
\centering
\includegraphics[width=\linewidth]{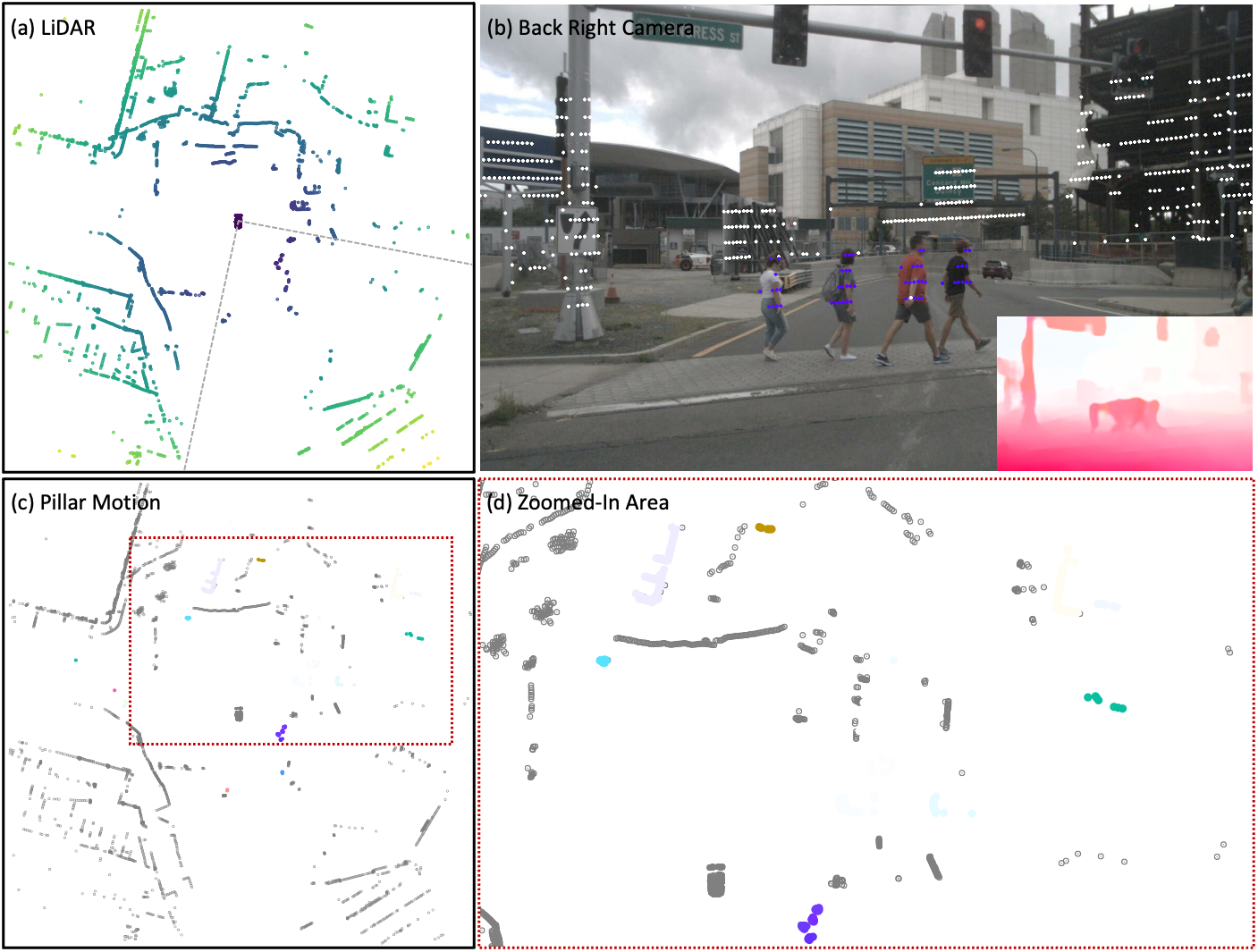}
\caption{An overview of the proposed self-supervised pillar motion learning by our designed free supervisory signals from point clouds and paired camera images. (a) illustrates a point cloud in BEV with the dotted gray lines showing the field of view of the back right camera. (b) shows the projected points with color encoding optical flow (ego-motion factorized out) on the back right camera image. Note that the white points are static. We attach the original optical flow of this image on bottom right for reference. (c) is the predicted pillar motion field, where hue and saturation correspond to motion direction and magnitude, and the gray are static pillars. (d) demonstrates a zoomed-in area of (c).}
\label{fig:teaser}
\vspace{-3mm}
\end{figure}

A traditional autonomy stack usually performs motion estimation by first recognizing other traffic participants in the scene and then predicting how the scene might progress given their current states~\cite{vector-net, lane-graph}. However, most recognition models are only trained to classify and localize objects from a handful of known categories. This closed-set scenario is apparently insufficient for a practical autonomy system to perceive motion of a large diversity of instances that are not seen during training. As the lower-level information compared to object semantics, motion should be ideally estimated in an open-set setting irrespective of whether objects belong to a known or unknown category. One appealing way to predict class-agnostic motion is to estimate scene flow from point clouds by estimating the 3D velocity of each point~\cite{flownet3d, just-go}. Unfortunately, this dense motion filed prediction is currently computationally prohibitive to process one complete LiDAR sweep, ruling out the practical use for self-driving vehicles that require real-time and large-scale point cloud processing.

Another possibility to represent and estimate motion is based on bird's eye view (BEV). In this way, a point cloud is discretized into grid cells, and motion information is described by encoding each cell with a 2D displacement vector indicating the position into the future of the cell on the ground plane~\cite{any-motion, pillar-flow, motion-net}. This compact representation successfully simplifies scene motion as the motion taking place on the ground plane is the primary concern for autonomous driving, while the motion in the vertical direction is not as much important or useful. Additionally, point clouds represented in this form are efficient since all key operations can be conducted via 2D convolutions that are extremely fast to compute on GPUs. Recent works also show that this representation can be readily generalized to class-agnostic motion estimation~\cite{any-motion, pillar-flow}. However, they have to rely on large amounts of annotated point cloud data with object detection and tracking as proxy motion supervision, which is expensive and difficult to obtain in practice. 

Statistics finds that a self-driving vehicle generates over 1 terabyte of data per day but only less than 5\% of the data is used~\cite{statistics}. Thus, learning without requiring manual labeling is of critical importance in order to fully harness the abundant data. While the recent years have seen growing interests in self-supervised learning for language~\cite{bert, albert} and vision~\cite{geometric, color}, self-supervision for point clouds still falls behind, yet has great potential to open up the possibility to utilize practically infinite training data that is continuously collected by the world-wide self-driving fleets.

\begin{figure*}[t]
\centering
\includegraphics[width=\linewidth]{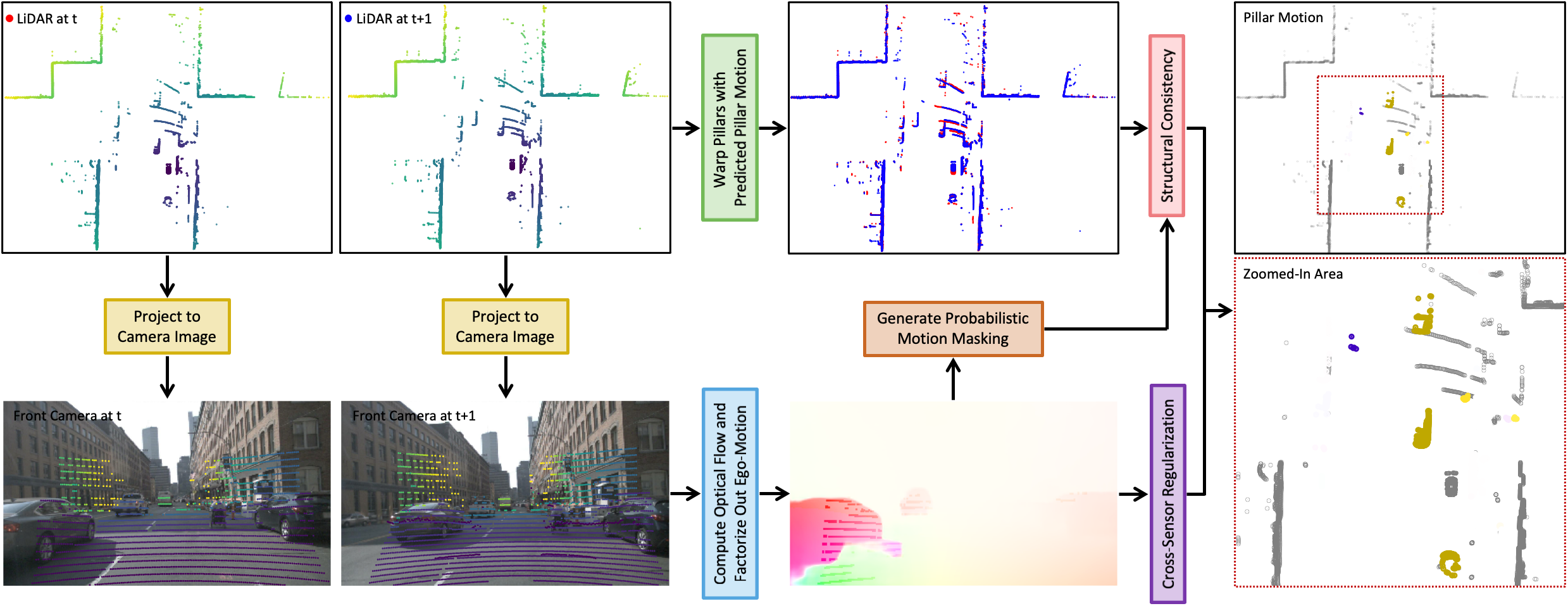}
\caption{A schematic overview of the proposed self-supervised learning framework for pillar motion estimation. We introduce a point cloud based structural consistency augmented with a probabilistic motion masking and a cross-sensor motion regularization to achieve the desired self-supervision. To illustrate the effect of factorizing out ego-motion, we plot such projected points on the original optical flow for comparison. Note that the bottom branch of the camera related components are discarded after training.}
\label{fig:method}
\vspace{-2mm}
\end{figure*}

In light of the above observations, we propose a self-supervised learning framework that exploits free supervisory signals from multiple sensors for open-set motion estimation, as shown in Figure~\ref{fig:teaser}. To take advantage of the merits of motion representation in BEV, we organize a point cloud into pillars (i.e., vertical columns)~\cite{pointpillars}, and refer to the velocity associated with each pillar as \textbf{pillar motion}. We introduce a point cloud based self-supervision by assuming pillar or object structure constancy between two consecutive sweeps. However, this does not hold in most cases due to the lack of exact point correspondence caused by the sparse scans of LiDAR. Our solution towards mitigating this difficulty is to make use of optical flow extracted from camera images to provide self-supervised and cross-sensory regularization. As illustrated in Figure~\ref{fig:method}, this design leads to a unified learning framework that subsumes the interactions between LiDAR and the paired cameras: (1) point clouds facilitate factorizing ego-motion out from optical flow; (2) image motion provides auxiliary regularization for learning pillar motion in point clouds; (3) probabilistic motion masking formed by back-projected optical flow promotes structural consistency in point clouds. Note that the camera-related components are only used in training and discarded for inference, thus no additional computations are introduced for the camera modality at runtime.   



To our knowledge, this work provides the first learning paradigm that is able to perform pillar motion prediction in a fully self-supervised framework. We propose novel self-supervisory and cross-sensory signals by tightly integrating point clouds and paired camera images to achieve the desired self-supervision. Experiments show that our approach compares favorably to the existing supervised methods. Our code and model will be made available at \url{https://github.com/qcraftai/pillar-motion}.



\section{Related Work}

\noindent \textbf{Motion Estimation.} This task aims to estimate motion dynamics and predict future locations of various agents via past observations. 
Traditional approaches typically formulate this task as a trajectory forecasting problem that hinges on perception outputs from 3D object detection and tracking~\cite{vector-net,lane-graph}. As a result of the dependence on the related modules, such a paradigm is prone to detection and tracking errors~\cite{pillar-flow} and lacks the ability to tackle unknown object classes~\cite{unknown}.
Another active research line is to estimate scene flow from point clouds to understand the dense 3D motion field~\cite{hpl-flownet, flownet3d}. However, current methods often take hundreds of milliseconds to process a partial point cloud, which is even though significantly subsampled. Moreover, the dense supervision of ground truth scene flow is hard to acquire in real data~\cite{just-go,point-pwc}. 
Thus, these methods operate on either synthetic data (FlyingThings3D~\cite{flying}) or densely processed data (KITTI Scene Flow~\cite{kitti}), where dense point correspondences are mostly available. However, for the raw point clouds scanned by LiDAR, such correspondences usually do not exist, making it more difficult to directly estimate scene flow from LiDAR. 

Several recent methods are proposed to explore estimating motion in BEV to simplify the understanding of scene motion. 
MotionNet is proposed in~\cite{motion-net} to perform joint perception and motion prediction based on a spatio-temporal pyramid network with three correlated heads. A differentiable ego-motion compensation layer~\cite{any-motion} is introduced to augment a recurrent convolutional network~\cite{prernn} for temporal context aggregation. In~\cite{pillar-flow} an optical flow network adapted from~\cite{pwc-cvpr} is used to learn correspondence matching between two consecutive point clouds. Compared to these works that rely on large labeled training data with 3D detection and tracking to approximate motion supervision, we aim to learn pillar motion from unlabeled data collections in a fully self-supervised manner.  

\noindent \textbf{Self-Supervised Learning.} To take advantage of the vast amount of unlabeled images and videos, numerous methods have been developed for different tasks using various self-supervised losses. For the feature representation learning, geometric or color transformation~\cite{geometric,color}, contrastive learning~\cite{hierarchy}, frame interpolation~\cite{interpolate-2}, and sequence ordering~\cite{ordering-1} have been widely explored. As for optical flow estimation~\cite{wang2018occlusion}, photometric constancy over time and spatial smoothness of flow fields are also well studied. In~\cite{autoencoder} the differentiable rendering and autoencoding reconstruction are applied to discover object attributes, such as 3D shape, landmarks, part segmentation, and camera viewpoint. 

All these studies are conducted in the context of images and videos, while self-supervision in point clouds has been less explored. PointContrast~\cite{point-contrast} presents a unified triplet of architecture and contrastive loss for pre-training, and transfers the learned 3D feature representations to point cloud segmentation and detection tasks. Mittal et al.~\cite{just-go} present a method to train scene flow by combining two self-supervised losses based on nearest neighbors and temporal consistency. It is proposed in~\cite{point-pwc} to further incorporate smoothness constraints and Laplacian coordinates to preserve local structure for scene flow training. Different from the previous works, we focus on pillar motion learning and leveraging complementary self-supervision from point clouds and associated camera images.

\noindent \textbf{LiDAR and Camera Fusion.} A large family of the multi-sensor fusion research is about 3D object detection. For object-centric methods~\cite{mv3d}, fusion is conducted at the object proposal level by roi-pooling features from separate backbone networks of camera and LiDAR. In~\cite{cont-fuse} continuous feature fusion is developed to allow feature sharing across all levels of the backbones of two modalities with a sophisticated mapping between point clouds and images. A detection seeding scheme is employed in~\cite{frustum} to extract image semantics from detection or segmentation to seed detection in point clouds. PointPainting~\cite{point-painting} presents a simple and sequential fusion method that projects points onto the output of an image based semantic segmentation network and appends the class scores to each point.         

Some recent works employ very sparse (hundreds to thousands) LiDAR measurements to enhance image based dense scene flow estimation. Sparse LiDAR is integrated with stereo images in~\cite{lidar-flow} to resolve the lack of information in challenging image regions caused by shadows, poor illumination, and textureless objects. Rishav et al.~\cite{deep-lidar-flow} further extend this method to a monocular camera setup through a late feature fusion at multiple scales and demonstrate superior performance over image-only methods. In contrast, we propose a cross-sensor based self-supervision to regularize motion learning in point clouds by alleviating the lack of exact correspondences between sweeps.

\section{Method}

As illustrated in Figure~\ref{fig:method}, the proposed motion learning approach tightly couples the self-supervised structural consistency from point clouds and the cross-sensor motion regularization. Our regularization involves factorizing out ego-motion from optical flow and enforcing motion agreement across sensors. We also introduce a probabilistic motion masking based on the back-projected optical flow to enhance structural similarity matching in point clouds.    


\subsection{Problem Formulation}

Given a temporal sequence of self-driving keyframes, we denote the point cloud and the paired camera images captured at time $t$ as $\mathcal{P}_t = \{P^t_i\}_{i=1}^{N_t}$ and $\mathcal{I}_t = \{I^t_i\}_{i=1}^{N_c}$, where $P^{t}_{i}$ indicates a point and $N_t$ is the number of received points, $I^t_i$ denotes an image and $N_c$ is the number of cameras in the sensor suite mounted on a self-driving vehicle. In the following we omit the camera index and use $I^t$ to indicate any camera image for brevity.  
$\mathcal{P}_t$ is discretized into non-overlapping pillars $\{\rho^t_i\}_{i=1}^{N_p}$, where $\rho^t_i$ denotes a pillar index and $N_p$ is the number of pillars.
We define the pillar motion field $\mathcal{M}_t = \{M^t_i\}_{i=1}^{N_p}$ as the movement of each pillar to its corresponding position at next timestamp:
$\tilde{\rho}^{t+1}_i = M^t_i(\rho^t_i)$, $M^t_i\in\mathbb{R}^2$. $\mathcal{M}_t$ is defined such that $\rho^t_i$ and $\tilde{\rho}^{t+1}_i$ represent the same part of a scene moving in time. $\mathcal{M}_t$ is a locally rigid and non-deforming motion field, which assumes that all dynamics are on ground plane and the motion is consistent for every point inside a pillar.

\subsection{LiDAR based Structural Consistency}

According to the above definition of pillar motion $\mathcal{M}_t$, we assign the motion vector $M^t_i$ of each pillar $\rho^t_i$ to all the points within the pillar to obtain the per point motion, and simply set the motion along the vertical direction to be zero. After having the per point motion, we can transform the original point cloud $\mathcal{P}_t$ to the next timestamp and get $\tilde{\mathcal{P}}_{t+1}$. Since there is no ground truth motion available, we instead to seek to use the structural consistency between the transformed point cloud $\tilde{\mathcal{P}}_{t+1}$ and the real point cloud $\mathcal{P}_{t+1}$ as free supervision to guide the pillar motion learning. 

We take inspiration from the chamfer matching for image registration~\cite{chamfer} and measure the structural similarity between two point clouds $\tilde{\mathcal{P}}$ and $\mathcal{P}$ as:
\begin{equation}
    \mathcal{L}_{\text{consist}} = \sum\limits_{\tilde{P}_i \in \tilde{\mathcal{P}}} \min\limits_{P_j \in \mathcal{P}} \|\tilde{P}_i - P_j\| + \sum\limits_{P_j \in \mathcal{P}} \min\limits_{\tilde{P}_i \in \tilde{\mathcal{P}}} \|P_j - \tilde{P}_i\|.
\label{eq:chamfer}
\end{equation}
Here we leave out the time index $t$ for brevity. 
Conceptually, this self-supervised structural consistency loss makes use of the nearest neighbor distances between transformed points and real points of the two point clouds $\tilde{\mathcal{P}}_{t+1}$ and $\mathcal{P}_{t+1}$ to approximate the pillar motion filed. 

Apparently this structural consistency relies on the existence of corresponding points between two consecutive point clouds $\mathcal{P}_t$ and $\mathcal{P}_{t+1}$. However, for the raw scans of LiDAR, this assumption does not usually hold for a number of cases. For example, the corresponding points cannot be exactly re-scanned at the next timestamp, and they can be occluded or fall out of the sensor range. These scenarios become even worse for the objects at distance, where the points are extremely sparse. As can be seen from the projected points on the two camera images in Figure~\ref{fig:method}, the scanned points on the back of the distant bus over the two sweeps do not correspond exactly. And the nearest neighbor matching can be ambiguous within cluster of points. Therefore, we argue that directly forcing the model to conduct the nearest neighbor based structural matching between two point clouds would inevitably introduce noise. 

 


\subsection{Cross-Sensor Motion Regularization}

As aforementioned, the nearest neighbor based structural consistency can cause ambiguity when point clouds are sufficiently sparse, as is common for the widely used sparse LiDAR. 
On the other hand, cameras in a sensor suite provide complementary appearance cues and more dense information. Thus, it would be helpful to incorporate image information along with LiDAR. One solution is to estimate scene flow from images also in an self-supervised manner. However, estimating scene flow directly from camera images is still hard and inaccurate, while optical flow estimation is relatively more precise and mature. So we instead relax the regularization by projecting predicted pillar motion onto image planes and utilize optical flow for the self-supervised motion regularization across sensors.          

However, it is infeasible to directly use optical flow to render additional motion supervision of a scene as optical flow is contaminated with both ego-motion and object motion. So we need to first factorize ego-motion out from optical flow. Let $F^t$ denote the optical flow estimated from two images $I^t$ and $I^{t+1}$, and at each pixel $(u,v)$ the optical flow can be decomposed into two parts:
\begin{equation}
    F^t(u,v) = F^t_{\text{ego}}(u,v) + F^t_{\text{obj}}(u,v),
\label{eq:flow-decompose}
\end{equation}
where $F^t_{\text{ego}}(u,v)$ is the motion caused by ego-vehicle motion, and $F^t_{\text{obj}}(u,v)$ is the true object motion. 
Given a point $P_{i}^{t}\in\mathbb{R}^3$ from $\mathcal{P}_t$ that is associated with $I^t$, the projected camera image location can be written as:
\begin{equation}
      (u_i, v_i) = KT_{\text{L}\rightarrow\text{C}}P^{t}_i,
\label{eq:project}
\end{equation}
where $K$ is the camera intrinsic parameters and $T_{\text{L}\rightarrow\text{C}}$ is the relative pose between LiDAR and the camera. We can then compute the optical flow part induced by ego-motion at location $(u_i, v_i)$ by:
\begin{equation}
     F^{t}_{\text{ego}}(u_i,v_i) = KT_{\text{L}\rightarrow\text{C}}T_{t\rightarrow t+1}P_{i}^t - KT_{\text{L}\rightarrow\text{C}}P_{i}^t,
\label{eq:flow-ego}
\end{equation}
where $T_{t\rightarrow t+1}$ is the ego-vehicle pose change. By combining Eqs.~\mref{eq:flow-decompose,eq:flow-ego}, we can factorize the ego-motion $F^t_{\text{ego}}$ out and obtain the pure object motion $F^t_{\text{obj}}$ to regularize the predicted pillar motion. Note that we only compute $F^t_{\text{obj}}$ at the pixels that have corresponding projected points as we can only compensate accurate $F^t_{\text{ego}}$ on these points. 

All the points within a pillar $\rho^t_i$ share the same predicted motion vector $M^t_i$. Again by using the projection function in Eq.~\mref{eq:project}, we can project the motion vector of each point onto the corresponding image plane and get the projected optical flow $\tilde{F}^t(u_i, v_i)$. By taking advantage of the projection relationship between point clouds and camera images, we establish the connection between pillar motion and optical flow, and thus enforce $\tilde{F}^t$ to be close to $F^t_{\text{obj}}$:
\begin{equation}
    \mathcal{L}_{\text{regular}}= \sum\limits_{i}\|\tilde{F}^t(u_i,v_i)-F^t_\text{obj}(u_i,v_i)\|_1.
\end{equation}

This cross-sensory loss serves as an auxiliary and important regularization to complement the structural consistency and mitigate the ambiguity due to large sparsity of point clouds. Furthermore, the optical flow guided regularization can be viewed as distilling motion knowledge from cameras to LiDAR during training. 
As for the optical flow estimation, we can train an optical flow model using an unsupervised method~\cite{wang2018occlusion} such that the whole learning framework enjoys an unified self-supervised setup. 


\begin{figure}[t]
\centering
\includegraphics[width=\linewidth]{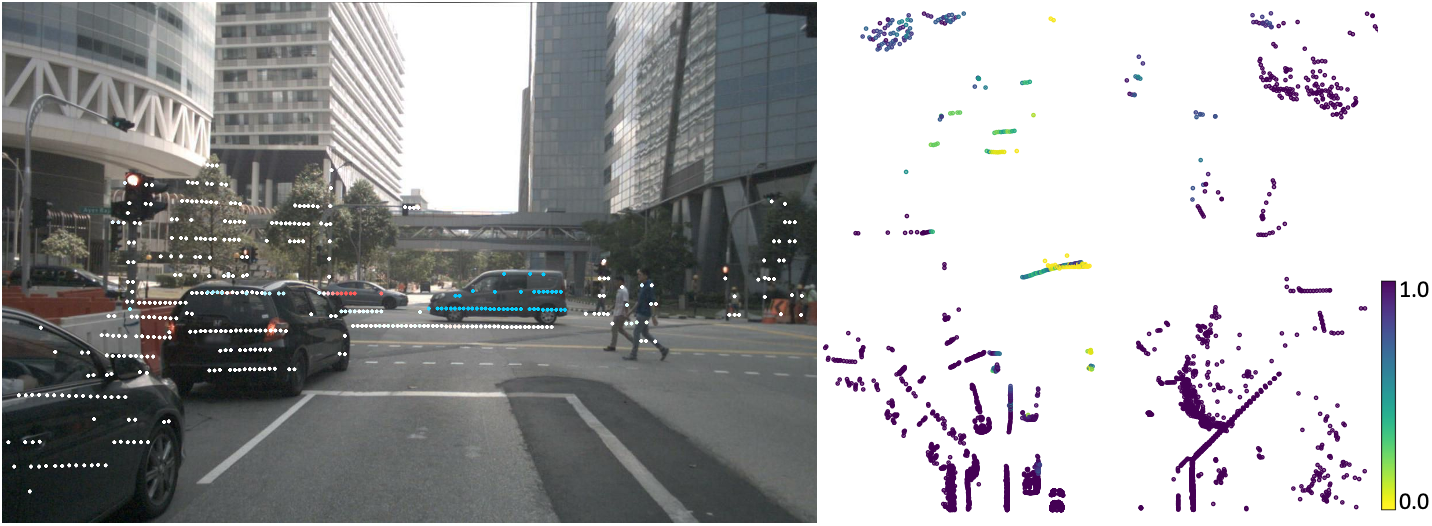}
\caption{Illustration of the probabilistic motion masking. Left: the projected points with color encoding optical flow (ego-motion factorized out) on the front image. Right: part of the point cloud and the probabilities of nonempty pillars being static.}
\label{fig:masking}
\vspace{-2mm}
\end{figure}

\subsection{Probabilistic Motion Masking}
\label{sec:masking}

After factorizing out the ego-motion, we can use the object motion part of optical flow to approximate a probabilistic motion mask to indicate the probability of each pillar being static or dynamic. Specifically, the probability of each projected point being static can be computed by: 
\begin{equation}
    s^t_i = \exp\{-\alpha\max(\|F^t_\text{obj}(u_i, v_i)\|-\tau, 0)\},
\label{eq:mask}
\end{equation}
where $\alpha$ is a smoothing factor and $\tau$ is a stationary tolerance. 
It is then back-projected to the point cloud coordinate, and we average $\{s^t_i\}$ of the points within a pillar to indicate the motion (being static) probability of this pillar, as shown in Figure~\ref{fig:masking}. When the ego-vehicle (LiDAR) is moving, the scanned points even from background and static foreground objects cannot be exactly re-scanned over time. This introduces noise to the static regions when enforcing the nearest neighbor matching in the structural consistency loss, which treats all points equally. We can leverage the probabilistic motion mask to downweight the points from static pillars. In practice, we can simply enhance Eq.~\mref{eq:chamfer} by adding a weighting coefficient that is represented as the pillar motion probability for each point. Furthermore, since static pillars are often dominant in a scene, this weighting strategy also helps to balance the contributions of static and dynamic pillars in computing the overall structural consistency loss.     



\subsection{Optimization}

Analogous to the spatial smoothness constraint for optical flow estimation, we also apply a local smoothness loss for pillar motion learning:
\begin{equation}
\mathcal{L}_{\text{smooth}}=|\nabla_x\mathcal{M}_t^x| + |\nabla_y\mathcal{M}_t^x| + |\nabla_x\mathcal{M}_t^y| + |\nabla_y\mathcal{M}_t^y|, 
\end{equation}
where $\mathcal{M}_t^x$ and $\mathcal{M}_t^y$ denote the $x$ and $y$ components of the predicted pillar motion field $\mathcal{M}_t$, and $\nabla_x$ and $\nabla_y$ are the gradients in $x$ and $y$ directions. Intuitively, this smoothness loss encourages the model to predict similar motion for the pillars belonging to the same object.   

In summary, the total loss is a weighted sum of three terms including the probabilistic motion masking weighted structural consistency loss, the cross-sensor motion regularization loss, and the local smoothness loss: 
\begin{equation}
    \mathcal{L}_{\text{total}}=\lambda_{\text{consist}}\mathcal{L}_{\text{consist}} + \lambda_{\text{regular}} \mathcal{L}_{\text{regular}} + \lambda_{\text{smooth}} \mathcal{L}_{\text{smooth}},
\label{eq:total}
\end{equation}
where $\lambda_{\text{consist}}$, $\lambda_{\text{regular}}$ and $\lambda_{\text{smooth}}$ are the balancing coefficients to control the importance of the three loss terms. We train our model to jointly optimize the total loss function.  

\subsection{Backbone Network}

Our proposed self-supervised learning framework is independent of the backbone network and can be generalized to various modified spatio-temporal networks~\cite{slow-fast, step}. In order to make fair comparisons with the existing supervised methods, we adopt a similar backbone network as~\cite{motion-net}. In details, we first employ a simple pillar feature encoder~\cite{pointpillars} that consists of a linear layer followed by batch normalization~\cite{batch-norm}, ReLU~\cite{relu} and a max pooling to convert a raw point cloud to a feature map representation in BEV. We then input the feature map to a U-Net with separated spatial and temporal convolutions as well as lateral connections between encoder and decoder.

\begin{table*}
\centering
\tabcolsep=0.07cm
\begin{tabular}{|cccc|cc|cc|cc|cc|cc|cc|}
\hline
  & \multirow{2}{*}{$\mathcal{L}_{\text{consist}}$} & \multirow{2}{*}{$\mathcal{L}_{\text{regular}}$} & \multirow{2}{*}{Mask} & \multicolumn{2}{c|}{Static} & \multicolumn{2}{c|}{Speed $\leq$ 5m/s} & \multicolumn{2}{c|}{Speed $>$ 5m/s} & \multicolumn{2}{c|}{Nonempty} & \multicolumn{2}{c|}{Foreground} & \multicolumn{2}{c|}{Moving} \\ \cline{5-16} 
 &   &                      &                       &                        Mean          & Median         & Mean          & Median         & Mean        & Median        & Mean         & Median         & Mean           & Median            & Mean          & Median        \\ \hline

(a) &  $\checkmark$  &   & &    0.3701          &      0.0063                 &      0.5014         & 0.1352 & 1.9405 &  1.2760 & 0.3437 & 0.0081 & 0.5936 & 0.1139 & 0.7516 &  0.2359  \\ \hline
(b) &  & $\checkmark$ &  & \textbf{0.0285} & \textbf{0.0002} & 0.3733 & \textbf{0.0719} & 4.2954 & 3.9788 & 0.0897 & 0.0020 & 0.7914 & 0.0656 & 1.1267 & 0.3948 \\ \hline
(c) &  $\checkmark$ &$\checkmark$  & &  0.1688 &  0.0389 & 0.4277 & 0.1694 & 1.7603 & 1.2021 & 0.3133  &  0.0062 &  0.5667  & 0.1017 &  0.7064 &  0.1980\\ \hline 
(d) &  $\checkmark$ & & $\checkmark$ & 0.0738 & 0.0038  & 0.4017  & 0.1214 &  1.9384 &  1.2931 &   0.1085  & 0.0007 & 0.5416 &  0.0767 & 0.8064  & 0.2279\\ \hline
(e) & $\checkmark$   & $\checkmark$ &  $\checkmark$ & 0.0619 & 0.0004  & \textbf{0.3438} & 0.1196 &  \textbf{1.7119} & \textbf{1.1438} & \textbf{0.0846} & \textbf{0.0001} &  \textbf{0.4494} & \textbf{0.0507} & \textbf{0.5953} & \textbf{0.1612}\\ \hline  
\end{tabular}
\vspace{4pt}
\caption{Comparison of our models using different combinations of the proposed structural consistency, cross-sensor regularization, and probabilistic motion masking. We evaluate each model on the six groups and report the mean and median errors.}
\label{tab:ablation}
\end{table*}

\section{Experiments}
In this section, we first describe our experimental setup. 
A variety of ablation studies are then performed to understand the contribution of each individual design in our approach. We report comparisons to the state-of-the-art methods on the benchmark dataset. In the end, we provide in-depth analysis with qualitative visualization results.

\subsection{Experimental Setup}
\noindent\textbf{Dataset.}
We extensively evaluate the proposed approach on a large-scale autonomous driving dataset: nuScenes~\cite{nuscenes}. It contains 850 scenes with annotations, and each scene is around 40s.  
Following the experimental protocol in MotionNet~\cite{motion-net}, we use 500 scenes for training, 100 scenes for validation and 250 scenes for testing. This dataset provides a full sensor suite including LiDAR, cameras, radars, IMU and GPS. We adopt LiDAR and all six cameras during training, and only use LiDAR for inference. 
The frequency of LiDAR and cameras are 20Hz and 10Hz, respectively.
We can derive the ground truth motion from the original detection and tracking annotations provided by the dataset.  
Unless specified, we only use the motion labeling in the evaluation stage, while only using the raw sensor inputs and the calibration data in the training stage.  

\begin{figure}[t]
\centering
\includegraphics[width=\linewidth]{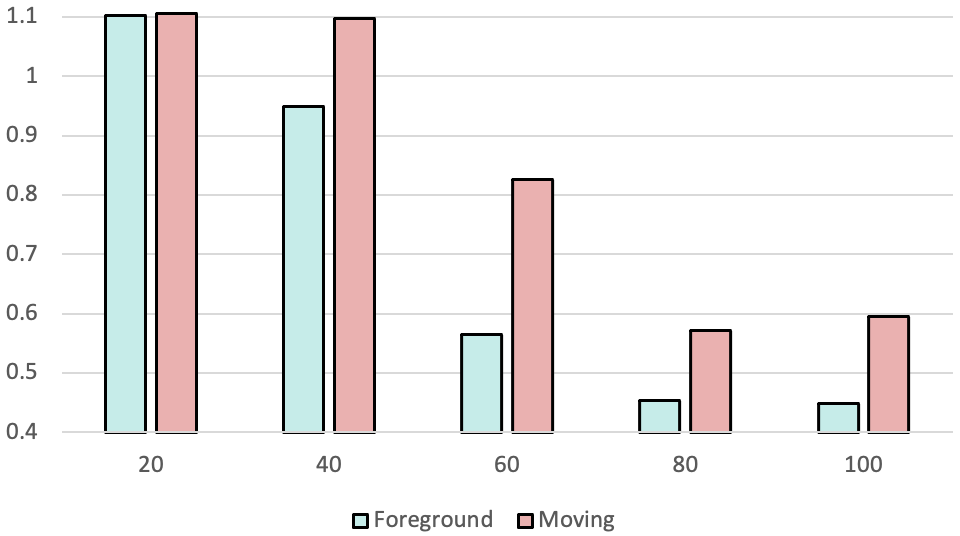}
\caption{Performance improvements of self-supervised motion estimation evaluated on the foreground and moving groups with increasing the fraction (\%) of total training data.}
\label{fig:amount}
\end{figure}

\noindent\textbf{Implementation Details.}
We implement our model in PyTorch~\cite{pytorch}, and train the model on 8 GPUs with a batch size of 64. We train the model for 200 epochs in total, and set the initial learning rate to be 0.0001 and decay the learning rate by a factor of 0.9 in every 20 epochs. AdamW~\cite{adamW} is used as the optimizer. 
We emperically set $\alpha = 0.1$ and $\tau = 5$ in Eq.~\mref{eq:mask}, and $\lambda_{\text{consist}}=1$, $\lambda_{\text{regular}}=0.01$ and $\lambda_{\text{smooth}}=1$ in Eq.~\mref{eq:total}. We take the current sweep and past four sweeps as input, and transform the four sweeps from past to the current coordinate system through ego-motion compensation. Our model outputs the displacement for the next 0.5s as the predicted motion. We follow~\cite{motion-net} to crop a point cloud in the range of $[-32\text{m}, 32\text{m}]\times[-32\text{m}, 32\text{m}]$, and set the pillar size to be $0.25\text{m}\times0.25\text{m}$ for fair comparisons.

We adopt PWC-Net~\cite{pwc} as the optical flow estimation network considering its accuracy and efficiency. It is unsupervised trained on the training images of nuScenes with occlusion-aware photometric constancy and spatial smoothness losses~\cite{wang2018occlusion}. We expect more advanced unsupervised methods such as ~\cite{luo2019every, wang2019unos} would improve optical flow quality and benefit our pillar motion learning further.




\noindent\textbf{Evaluation Metrics.}
Following MotionNet~\cite{motion-net}, we report the mean and median errors measured on the nonempty pillars, which are divided into three speed groups, i.e., static, slow ($\leq 5$m/s) and fast ($> 5$m/s). Additionally, we also evaluate on all nonempty pillars, all foreground object pillars, and all moving object pillars. The reason for evaluating on the different groups is that the static regions in a scene are the majority, which would overwhelm the prediction error if averaging over the whole scene. 


\subsection{Ablation Studies}
\noindent\textbf{Contribution of Individual Component.} We first perform a variety of combination experiments to evaluate the contribution of each individual component in our design. As shown in Table~\ref{tab:ablation}, the base model (a) trained only with the structural consistency loss does not perform well for the static group. This is in accordance with our previous analysis.   
By using the cross-sensor motion regularization as the only supervision, model (b) achieves impressive result for the static group thanks to the accurate ego-motion factorization that enables our approach to reliably recover the static points from optical flow. However, the result of this model for the fast speed group is far inferior. This is not surprising since merely having the motion regularization in 2D camera image space is ambiguous and there exists numerous possible motion in 3D point cloud space that corresponds to the same 2D projection. Above limitations of the models (a, b) together demonstrate the necessity to have motion self-supervision in both 2D and 3D.       


Although model (c) that directly combines the structural consistency and motion regularization performs well for the fast speed group, it is still not optimal for the static and slow speed groups, mainly due to the inconsistency between the two losses on the static and slow moving regions. 
By integrating the probabilistic motion masking into model (c), the full model (e) achieves significant improvements for static and slow speed groups. This is because by suppressing the plausible static pillars, the model can be less confused by the noisy motion caused by the moving ego-vehicle, and therefore can better focus on learning of the true object motion. 
We also experiment with model (d) that is trained only with the probabilistic motion masking enhanced structural consistency. Compared to model (a), the improvements on static and slow speed groups are obvious. However, it is still inferior to the full model (e), which further validates the efficacy of the cross-sensor motion regularization to provide the complementary motion supervision.  

\begin{table*}[h]
\centering
\begin{tabular}{|l|c|c|c|c|c|c|c|c|c|}
\hline
\multirow{2}{*}{Method} & \multicolumn{3}{c|}{Static}                                                             & \multicolumn{3}{c|}{Speed $\leq$ 5m/s}                                                            & \multicolumn{3}{c|}{Speed $>$ 5m/s}                                                           \\ \cline{2-10} 
                  & \multicolumn{1}{l|}{$<$ 10m} & \multicolumn{1}{l|}{$<$ 20m} & \multicolumn{1}{l|}{$<$ 30m} & \multicolumn{1}{l|}{$<$ 10m} & \multicolumn{1}{l|}{$<$ 20m} & \multicolumn{1}{l|}{$<$ 30m} & \multicolumn{1}{l|}{$<$ 10m} & \multicolumn{1}{l|}{$<$ 20m} & \multicolumn{1}{l|}{$<$ 30m} \\ \hline
$\mathcal{L}_{\text{consist}}$          & 0.2042                      & 0.2296                      & 0.3701                      & 0.4343                      & 0.4190                      & 0.5014                      & 1.7702                      & 1.8729                      & 1.9405                      \\ \hline
$\mathcal{L}_{\text{consist}}$ + $\mathcal{L}_{\text{regular}}$ & 0.1424 & 0.1602 & 0.1688  & 0.4001 & 0.4167 & 0.4277 & 1.7599  &  1.7576 & 1.7603\\
\hline
Full Model           &       \textbf{0.1036}                &     \textbf{0.0857}                   & \textbf{0.0619}                     &      \textbf{0.3348}                 &     \textbf{0.3412}              & \textbf{0.3438}                    & \textbf{1.6904}                     &    \textbf{1.6998}               & \textbf{1.7119}                    \\ \hline
\end{tabular}
\vspace{5pt}
\caption{Comparison of motion estimation results of our approach using the structural consistency, its combination with cross-sensor regularization, and the full model within different ranges of LiDAR. We report the mean error on the three speed groups.}
\label{tab:dist}
\end{table*}

\begin{table*}[h]
\centering
\begin{tabular}{|l|c|c|c|c|c|c|c|}
\hline
\multirow{2}{*}{Method} & \multicolumn{2}{c|}{Static} & \multicolumn{2}{c|}{Speed $\leq$ 5m/s} & \multicolumn{2}{c|}{Speed $>$ 5m/s} & \multirow{2}{*}{Time} \\ \cline{2-7} 
                        & Mean        & Median        & Mean           & Median           & Mean           & Median    &       \\ \hline
FlowNet3D (pre-trained)~\cite{flownet3d}  & 2.0514 & \textbf{0.0000} & 2.2058 & 0.3172 & 9.1923    & 8.4923 & 0.434s     \\ \hline
HPLFlowNet (pre-trained)~\cite{hpl-flownet} & 2.2165 & 1.4925 & 1.5477 & 1.1269 & 5.9841 & 4.8553 & 0.352s \\\hline
\textbf{Ours (self-supervised)} & \textbf{0.1620} & 0.0010 & \textbf{0.6972} & \textbf{0.1758} & \textbf{3.5504} & \textbf{2.0844} & 0.020s \\ \hline \hline
FlowNet3D~\cite{flownet3d} & 0.0410 & 0.0000 & 0.8183 & 0.1782 & 8.5261 & 8.0230 & 0.434s \\ \hline
HPLFlowNet~\cite{hpl-flownet} & 0.0041 & 0.0002 & 0.4458 & 0.0969 & 4.3206 & 2.4881 & 0.352s \\ \hline
PointRCNN~\cite{point-rcnn} & 0.0204 & 0.0000 & 0.5514 & 0.1627 & 3.9888 & 1.6252 & 0.201s \\ \hline
LSTMEncoderDecoder~\cite{lstm-encoder-decoder} & 0.0358 & 0.0000 & 0.3551 & 0.1044 & 1.5885 & 1.0003 & 0.042s \\ \hline

MotionNet~\cite{motion-net} & 0.0239 & 0.0000 & 0.2467 & 0.0961 & 1.0109 & 0.6994 & 0.019s \\ \hline
MotionNet (pillar-based)~\cite{motion-net} & 0.0258 & - &  0.2612 & - & 1.0747 & - & 0.019s \\ \hline
MotionNet+MGDA~\cite{motion-net} & \textbf{0.0201} & 0.0000 & 0.2292 & 0.0952 & 0.9454 & 0.6180 & 0.019s \\ \hline 
\textbf{Ours (fine-tuned)} &  0.0245 &  \textbf{0.0000}  & \textbf{0.2286} & \textbf{0.0930} & \textbf{0.7784}  & \textbf{0.4685} & 0.020s  \\ \hline
\end{tabular}
\vspace{5pt}
\caption{Comparison with the state-of-the-art results. We report the mean and median errors on the three speed groups. Top: we compare our self-supervised model to the methods that are not trained with the annotations of nuScenes but are supervised pre-trained on two scene flow datasets. Bottom: we fine-tune our self-supervised model and compare to the methods that are supervised trained on nuScenes.}
\label{tab:sota}
\end{table*}


\noindent \textbf{Amount of Training Data.}
Next we study how the amount of training data impacts our self-supervised learning. Since the foreground and moving objects are more concerned and challenging, we plot the mean errors of our full model on the two groups under different percentages of training data in Figure~\ref{fig:amount}. With the increase of training size from $20\%$ to $100\%$, our approach can boost the performance from $1.1021$ to $0.4494$ and $1.1067$ to $0.5953$ for the foreground and moving groups. Overall, more unlabeled training data leads to significantly better prediction results, suggesting the great potential of our self-supervised approach to harvest the vast amount of self-driving data that is available today. 

\noindent \textbf{Performance vs. Distance.} Since the point clouds become more and more sparse with the increasing of distance, we also study how our model behaves at different ranges of LiDAR. As shown in Table~\ref{tab:dist}, we compute the mean errors of each speed group within the distances of 10m, 20m and 30m, respectively. Specifically, the performance of our full model degrades $0.0417$, $0.0090$ and $0.0215$ from near to far regions in the three speed groups, as compared to the much larger performance drops of $0.1659$, $0.0824$ and $0.1703$ for the structural consistency only based model. Combining structural consistency and cross-sensor regularization also achieves apparent improvement over the base model. This verifies that optical flow provides more dense information complementary to point clouds and our full model regularized by optical flow is more robust to the distant pillars.       

\subsection{Comparison with State-of-the-Art Results}
As reported in the ablation studies, our model is trained to predict the displacements for next 0.5s as the keyframes are sampled at 2Hz in nuScenes.  
Here for fair comparisons with the methods that evaluate for next 1.0s, we simply linearly interpolate our predicted displacements to 1.0s by assuming constant velocity in a short time frame.

\begin{figure*}[t]
\centering
\includegraphics[width=0.99\linewidth]{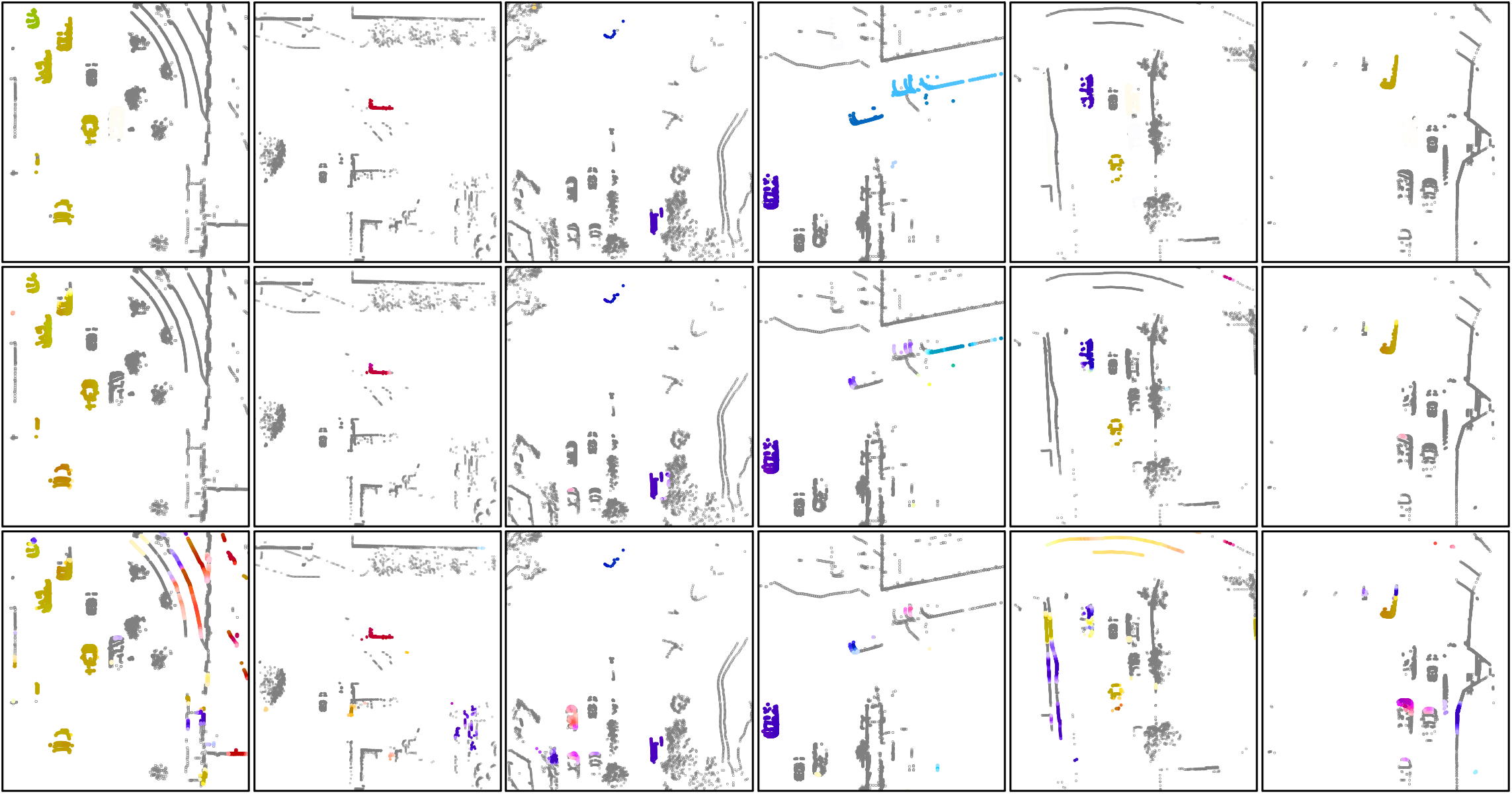}
\caption{Comparison of the predicted pillar motion. We show the ground truth motion field in the first row, the results estimated by our full model in the second row, and the predictions by the base model using only structural consistency in the third row. Each column demonstrates one scene. We remove the ground points for better visualization. }
\label{fig:qualitative}
\end{figure*}

We extensively compare our approach with a variety of supervised algorithms in Table~\ref{tab:sota}. For our self-supervised model, we first compare to FlowNet3D and HPLFlowNet, which are pre-trained on FlyingThings3D and KITTI Scene Flow. As can be seen in this table, our model largely outperforms the two methods that are supervised pre-trained though. Remarkably, our self-supervised model is found to even outperform or approach some methods that are fully supervised trained on the benchmark dataset, for instance, our model performs better than FlowNet3D, HPLFlowNet and PointRCNN for the fast speed group. All these comparisons collectively and clearly show the advantage of our proposed self-supervisory design and the importance of self-supervised training on the target domain.         

When further fine-tuning our self-supervised model with the ground truth labels, our approach achieves the state-of-the-art result. As we can see in Table~\ref{tab:sota}, our fine-tuned model clearly outperforms the related methods of MotionNet for the fast moving objects. In particular, when compared to the strong MotionNet+MGDA, we perform better with a clear margin of $0.1670$ mean error and $0.1495$ median error for the fast speed group. This indicates that our self-supervised model provides a better foundation to allow for more effective supervised training, and the self-supervised learning gain does not diminish with the sophisticated supervised training design.  


\subsection{Runtime Analysis}
At the inference stage, our whole model runs at 20ms on a single TITAN RTX GPU. In more details, the point cloud transformation and voxelization use 10ms, and the network forward time takes 10ms. As shown in Table~\ref{tab:sota}, compared with the existing point cloud based motion estimation networks, our model is more computational efficient and is capable of dealing with large-scale point clouds in real time.  

\subsection{Qualitative Results}
Finally, we demonstrate the qualitative results of pillar motion estimation using different combinations of the proposed self-supervised components. As shown in Figure~\ref{fig:qualitative}, these examples present diverse traffic scenes and different zoom-in scales. In comparison to our full model, the base model using only structural consistency loss tends to generate false positive motion predictions in the background regions (columns 1 and 5) and static foreground objects (columns 2 and 3). This observation verifies our interpretation in Section~\ref{sec:masking} that the noise induced by the moving ego-vehicle is detrimental to the structural consistency matching when applied in the background and static foreground pillars. Our full model can successfully eliminate most false positive motion, indicating that the optical flow based regularization and masking are effective to suppress such noise. Compared to the base model, the full model is also able to produce spatially smoother motion on the moving objects (columns 5 and 6). Moreover, as illustrated in column 4, a moving truck on top right of the scene is missing in the base model, but it is reasonably well estimated by our full model. This again validates the efficacy of the distilled motion information from camera images. 


\section{Conclusion}

In this paper, we propose a self-supervised learning framework for pillar motion estimation from unlabeled collections of point clouds and paired camera images. Our model involves a point cloud based structural consistency that is augmented with probabilistic motion masking and a cross-sensor motion regularization. Extensive experiments demonstrate that our self-supervised approach achieves superior or comparable results to the supervised methods, and outperforms the state-of-the-art methods when our model is further supervised fine-tuned. We hope these findings would encourage more research works on pillar motion estimation and point cloud based self-supervised learning.


\appendix
\input{appendix.tex}

\clearpage

\newpage
{\small
\bibliographystyle{ieee_fullname}
\bibliography{egbib}
}

\end{document}

%% file: appendix.tex
\section*{Appendix}



\section{Self-Supervision for Supervised Training}
\label{sec:amount}
Our self-supervised learning can be utilized as unsupervised pre-training to improve supervised training. In this section, we investigate the benefits of our self-supervised pre-trained models to supervised training under different amounts of training data with the derived motion annotations. Specifically, we randomly sample 20\%, 40\%, 60\% and 80\% of the entire training data. We compare the models trained from scratch against the ones initialized from the self-supervised pre-trained models. Here we summarize the important findings from the comparisons reported in Table~\ref{tab:amount}. (1) It is observed that the self-supervised pre-trained models consistently and significantly outperform the randomly initialized models across all cases. Such improvements are more remarkable under fewer training data and for the fast speed group. (2) Our self-supervised model fine-tuned with a small amount of training data (i.e., 20\%) is able to achieve comparable performance compared to the randomly initialized model trained with a large amount of training data (i.e., 80\%). This suggests that the model with self-supervised pre-training requires much fewer annotations and is therefore more labeling efficient. (3) We find that the self-supervised pre-trained models converge faster, taking about 60\% of the training iterations as the models initialized from scratch.     


\section{Pillar Motion for Downstream Tasks}
\label{sec:downstream}
\begin{figure}[t]
\centering
\includegraphics[width=\linewidth]{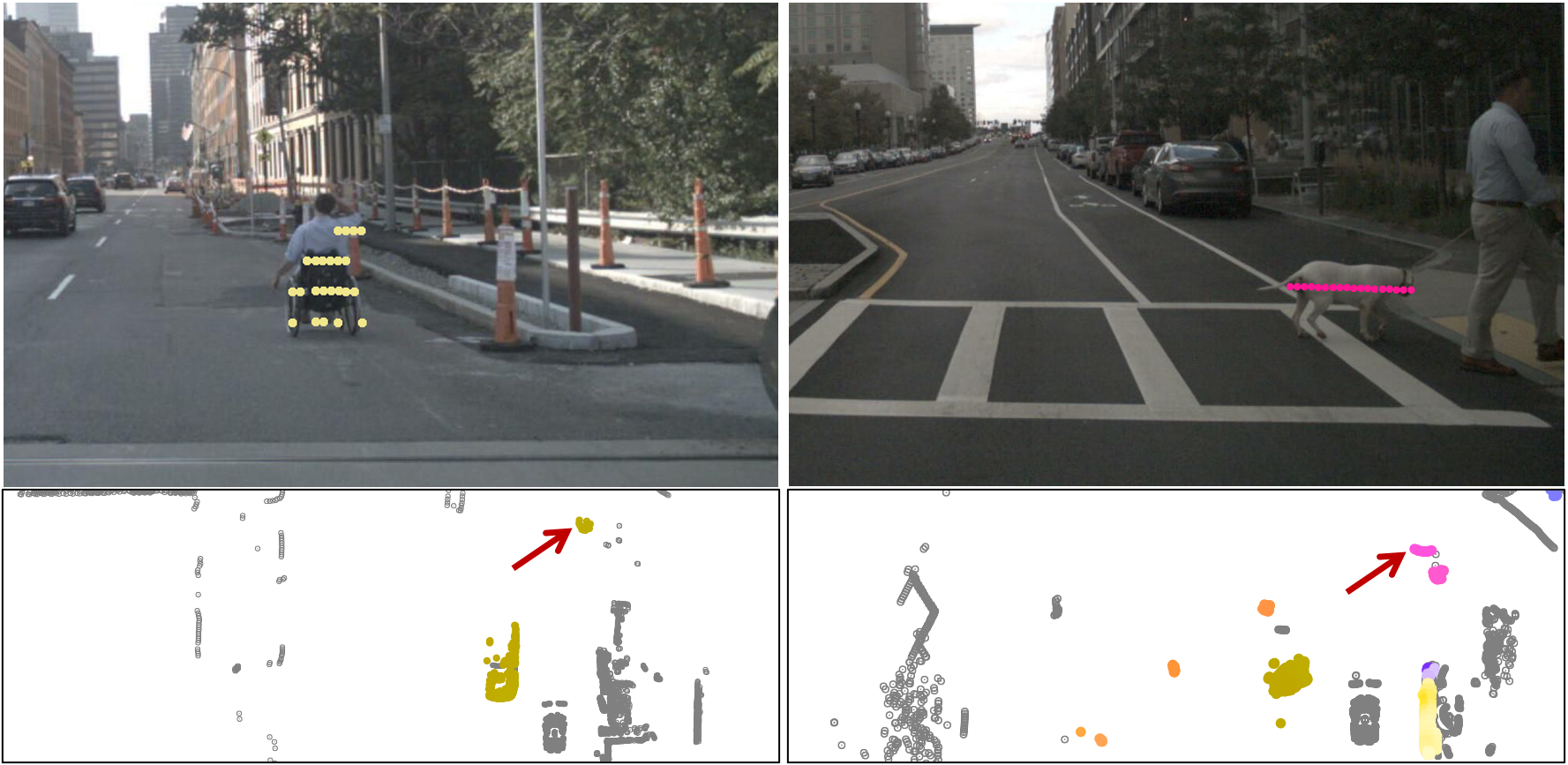}
\caption{Examples of perceiving the rare objects: wheelchair and dog, which are not seen in the training of point cloud based 3D object detection. We show the results (indicated by the two red arrows) of our self-supervised model, which can correctly estimate the class-agnostic pillar motion.}
\label{fig:agnostic}
\end{figure}

Our pillar motion can be potentially applied to enhance a variety of downstream modules. For perception, we show one advantage of our model to deal with the unknown instances that are not seen during the training of 3D object detection, as illustrated in Figure~\ref{fig:agnostic}. For tracking, it is empirically demonstrated in~\cite{pillar-flow} that integrating the low-level motion information improves the object tracking performance. As for planning, knowing the moving state of an agent is particularly helpful to tackle rare objects although the specific classes are unknown.

\begin{table*}[t]
\centering
\begin{tabular}{|cc|c|c|c|c|c|c|}
\hline
\multirow{2}{*}{Amount} & \multirow{2}{*}{Self-Supervised} & \multicolumn{2}{c|}{Static} & \multicolumn{2}{c|}{Speed $\leq$ 5m/s} & \multicolumn{2}{c|}{Speed $>$ 5m/s} \\ \cline{3-8} 
                   &      & Mean        & Median        & Mean           & Median           & Mean           & Median           \\ \hline
0\% & \cmark & 0.1620 & 0.0010 & 0.6972 & 0.1758 & 3.5504 & 2.0844 \\ \hline
\multirow{2}{*}{20\%} & \xmark &  0.0473 & 0.0001 & 0.4635 & 0.1400 & 2.0946 & 1.1676\\ \cline{3-8}
   & \cmark & 0.0394 & 0.0001 & 0.2970 & 0.1309 & 1.028 & 0.6055 \\ \hline 
\multirow{2}{*}{40\%} & \xmark &  0.0459 & 0.0001 & 0.3712 & 0.1385 & 1.7060 & 0.8950 \\ \cline{3-8}
 & \cmark & 0.0329 & 0.0000 & 0.2813 & 0.1280 & 0.8923 & 0.5287 \\ \hline
\multirow{2}{*}{60\%} & \xmark &  0.0412 & 0.0001 & 0.3082 & 0.1338 & 1.0912 & 0.6830\\ \cline{3-8}
 &  \cmark & 0.0352 & 0.0000 & 0.2801 & 0.1297 & 0.8499 & 0.5148   \\ \hline
\multirow{2}{*}{80\%} & \xmark &  0.0347 & 0.0001 & 0.2930 & 0.1322 & 0.9824 & 0.6110\\ \cline{3-8}
& \cmark &0.0247 & 0.0000 & 0.2301 & 0.0933 & 0.7788 & 0.4700  \\ \hline
\end{tabular}
\vspace{6pt}
\caption{Benefits of our self-supervised pre-training under different amounts of training data. \cmark: the models are first self-supervised pre-trained and then supervised fine-tuned with the annotations of nuScenes. \xmark: the models are randomly initialized from scratch and supervised trained with the annotations of nuScenes. We report the mean and median errors on the three speed groups. Note: no fine-tuning is performed under 0\%, which is provided as a baseline reference. }
\label{tab:amount}
\end{table*}

\vspace{4pt}

\begin{figure*}[t]
\centering
\includegraphics[width=\linewidth]{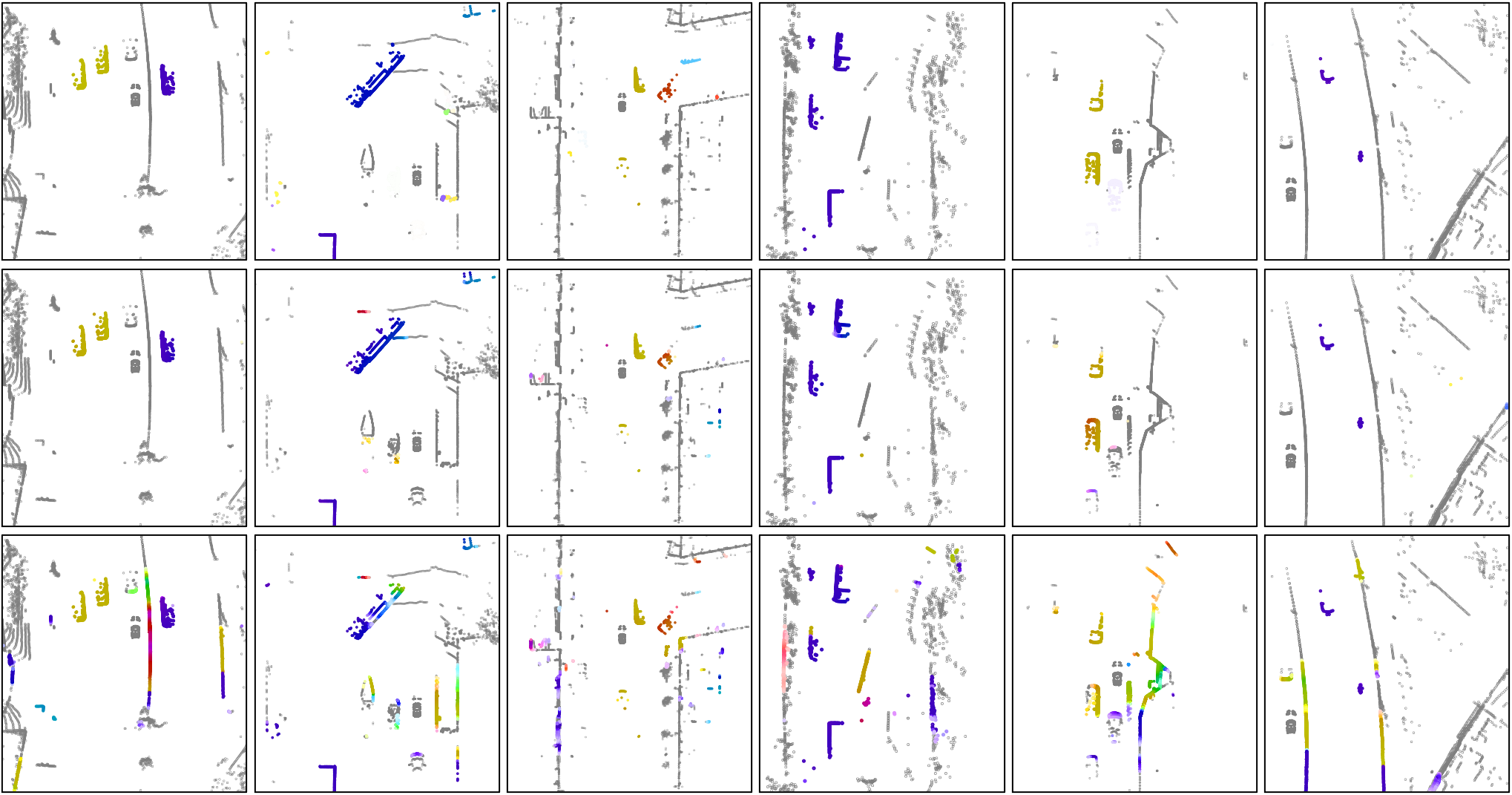}
\caption{Comparison of the predicted pillar motion. We show the ground truth motion field in the first row, the results estimated by our full model in the second row, and the predictions by the model without using probabilistic motion masking in the third row. Each column demonstrates one scene. We remove the ground points for better visualization.}
\label{fig:qualitative-supp}
\end{figure*}

\section{Comparison with Scene Flow Estimation}
\label{sec:compare}
In addition to the comparisons described in the introduction and related work of the paper, here we elaborate more on the differences between our work and the previous methods for point cloud based scene flow estimation. First, scene flow aims to estimate the point correspondences between two point clouds, while our goal is to predict the motion of each pillar or the displacement vector that indicates the future position of each pillar.
Second, although apart from the synthetic data (e.g., FlyingThings3D), the existing scene flow methods also experiment with the self-driving data (e.g., KITTI Scene Flow), they do not use the raw LiDAR scans. Instead, they combine 2D optical flow with depth map and convert them into 3D scene flow. Compared with the point clouds collected by LiDAR, the converted point clouds are much more dense. However, for the raw point clouds used by self-driving vehicles, this does not hold in most cases, making the task harder, in particular for directly doing self-supervision. Third, the prior scene flow methods usually take hundreds of milliseconds when operating on a partial point cloud that is even largely subsampled. Our approach can achieve pillar motion prediction of a complete point cloud in real-time. 

\section{Ablation Study on Smoothness}
\label{sec:smooth}
Removing the smoothness term in Eq.~\ref{eq:total} slightly increases the mean errors, e.g., 0.0058 (Speed $\leq$ 5m/s), 0.0041 (Speed $>$ 5m/s), and 0.0042 (Moving). 
Overall, the smoothness loss in pillar motion is not as significant as in optical flow. This is due to the fact that the form of pillar motion representation already implies the smoothness prior as each pillar shares the same motion in $0.25\text{m}\times0.25\text{m}$. In addition, the motion prediction of empty pillars that occupy a large portion of areas can be directly masked out. 

\section{Hyper-Parameters}
\label{sec:hyper}
We set the hyper-parameters to roughly balance the different loss terms: $\lambda_{\text{consist}} = \lambda_{\text{smooth}} = 1$ and $\lambda_{\text{regular}} = 0.01$. We also experiment with $\lambda_{\text{regular}} = 0.02, 0.03, 0.04, 0.05$. Under the five values of $\lambda_{\text{regular}}$, the standard deviations of mean errors of the three speed groups are very low: 0.0004, 0.0010 and 0.0070, indicating the robustness of our model to the hyper-parameter setting.

\section{More Qualitative Results}
\label{sec:qualitative}

Next we provide more qualitative results to reveal the efficacy of the proposed probabilistic motion masking. In Figure~\ref{fig:qualitative-supp}, we compare the predicted pillar motion fields by our full model and the model without using probabilistic motion masking. As shown in this figure, we present 6 scenes with diverse traffic scenarios and multiple zoom-in scales. In comparison to our full model, the model not using probabilistic motion masking tends to produce more false positive motion predictions at the background regions, such as building, wall and vegetation. This comparison further validates the effect of probabilistic motion masking to reduce the noise incurred by the moving ego-vehicle to the pillars of the background regions.